# Combining Deep Reinforcement Learning and Safety Based Control for Autonomous Driving

Xi Xiong    Jianqiang Wang    Fang Zhang    Keqiang Li

State Key Laboratory of Automotive Safety and Energy, Tsinghua University

**Abstract**

With the development of state-of-art deep reinforcement learning, we can efficiently tackle continuous control problems. But the deep reinforcement learning method for continuous control is based on historical data, which would make unpredicted decisions in unfamiliar scenarios. Combining deep reinforcement learning and safety based control can get good performance for self-driving and collision avoidance. In this passage, we use the Deep Deterministic Policy Gradient algorithm to implement autonomous driving without vehicles around. The vehicle can learn the driving policy in a stable and familiar environment, which is efficient and reliable. Then we use the artificial potential field to design collision avoidance algorithm with vehicles around. The path tracking method is also taken into consideration. The combination of deep reinforcement learning and safety based control performs well in most scenarios.

**1. Introduction**

There are two major paradigms for autonomous driving: the learning method and the control method. With the success of deep learning and reinforcement learning, more and more people have focused on using the learning method for autonomous driving. The combination of deep learning and reinforcement learning can tackle problems of high dimensional inputs. The DQN network (Minh et al., 2015) can play Atari games at the human level. But the DQN is not efficient to solve problems with high dimensional action state spaces. The combination of Q-network with actor-critic structure can perform well in the continuous control field. The DDPG algorithm (Lillicrap et al., 2015) presents an actor-critic, model-free algorithm based on the deterministic policy gradient that can operate over continuous action spaces. This algorithm can learn policies end-to-end: from low-dimension inputs or raw pixel inputs to final actions.

In the DDPG algorithm, the positive reward is the velocity projected along the track direction. We want the vehicle to run along the track as fast as possible. The negative reward is the penalty for collision. However, this method can't perform well without sufficient training. The policy would make unpredicted decisions in unfamiliar scenarios, which is also the shortcoming of data based method. In addition, avoiding collision is the basic function when designing the control strategy for autonomous driving. We also want the vehicle to run on the road with higher safety level, including driving along the central road track and keeping safety distances from vehicles around.

The artificial potential field method is widely used for collision avoidance in the field of robot path planning. We combine the ideas in the artificial potential field (Khatib, 1986) with deep reinforcement learning for autonomous driving to put both merits into full use.

Path tracking is also important for the autonomous driving strategy because we assume that driving far away from the center of the road is with high risk. We can use the path tracking method to get a relative safe state.

## 2. Background

### 2.1. Deep Reinforcement Learning

In the structure of reinforcement learning, the agent interacts with the environment. After every discrete time $t$, the agent implements the action $a$. Then, the environment changes its previous state $s_t$ to $s_{t+1}$, and the agent gets its reward $r_t$. The goal of reinforcement learning is to maximize the discounted accumulative reward $R_t = \sum_{i=t}^{T} \gamma^{i-t} r(s_i, a_i)$. The action-value function is used to express the expectation of $R_t$. $Q(s_t, a_t) = E(R_t | s_t, a_t)$. We define the optimal action value function as the maximum achievable expected return under the strategy $\pi$,

$$Q^*(s,a) = \max_{\pi} E\left[ R_t | s_t = s, a_t = a, \pi \right] \tag{1}$$

The optimal value functions can be expressed with Bellman equations,

$$Q^*(s,a) = E\left[ r + \gamma \max_{a_{t+1}} Q^*(s_{t+1}, a_{t+1}) | s, a \right]. \tag{2}$$

In DQN algorithm, it's common to use neural networks to approximate the $Q^*(s,a)$. We assume the approximator as $Q(s,a;\theta) \approx Q^*(s,a)$, with the parameters $\theta$. The Q-network can be trained by minimizing the loss function,

$$L_i(\theta_i) = E\left[ (r + \gamma \max_{a_{i+1}} Q(s_{i+1}, a_{i+1}; \theta_{i-1}) - Q(s_i, a_i; \theta_i))^2 \right] \tag{3}$$

The DQN algorithm can work well in high dimensional state spaces but is not effective in continuous action spaces because the optimization of $a_t$ at every time step is too slow to be practical with nontrivial action spaces. We use the Deep Deterministic Policy Gradient algorithm to solve continuous control problems.

We divide the control policies into the stochastic policy and the deterministic policy (Sutton et al., 2012). We assume the stochastic policy as $\pi(a|s) = P(a|s)$, which represents the action probability distribution. We also denote the state distribution as $\rho^{\pi}(s)$. The objective performance can be expressed as an expectation,

$$J(\pi_\theta) = \int_S \rho^{\pi}(s) \int_A \pi_\theta(a|s) \cdot Q^{\pi}(s,a) \mathrm{d}a \mathrm{d}s \tag{4}$$

The essence of policy gradient algorithm is to adjust the parameters of the policy in the direction of the performance gradient $\nabla_\theta J(\pi_\theta)$. The *policy gradient theorem* (Sutton et al, 1999) can be expressed as,

$$\nabla_\theta J(\pi_\theta) = \int_S \rho^\pi(s) \int_A \nabla_\theta \pi_\theta(a|s) \cdot Q^\pi(s,a) \mathrm{d}a \mathrm{d}s$$

$$= E\left[ Q^\pi(s,a) \cdot \nabla_\theta \ln \pi_\theta(a|s) \right] \qquad (5)$$

As for the continuous control problems, we assume the policy to be deterministic. We use $\mu_\theta$ to represent the reflection from the state spaces to the action spaces, namely $a = \mu_\theta(s)$. As with the definition of stochastic policy, we define the objective performance as,

$$J(\mu_\theta) = \int_S \rho^\mu(s) \cdot Q^\mu(s,a) \mathrm{d}s = \int_S \rho^\mu(s) \cdot Q^\mu(s, \mu_\theta(s)) \mathrm{d}s \qquad (6)$$

We also use the policy gradient for the deterministic policy. If $\nabla_\theta \mu_\theta(s)$ and $\nabla_a Q^\mu(s,a)$ both exist, then the gradient can be expressed as,

$$\nabla_\theta J(\mu_\theta) = \int_S \rho^\mu(s) \cdot \nabla_\theta \mu_\theta(s) \cdot \nabla_a Q^\mu(s,a) \mathrm{d}s = E\left[ \nabla_a Q^\mu(s,a) \cdot \nabla_\theta \mu_\theta(s) \right] \qquad (7)$$

**2.2. Safety Based Control**

When considering the safety of the vehicle, avoiding collision and driving along the track are the most important issues, especially the former function.

The artificial potential field method is widely used for robot path planning. The goal of potential field method is to make the robot move from the initial position to the target position in a desired manner while avoiding collision.

There are two types of potential field in the domain of robot path planning, the attractive potential field and repulsive potential field. The attractive potential part represents the energy to get to the target position. The repulsive potential part represents the potential risk of collision.

$$U_{art}(x) = U_{att}(x) + U_{rep}(x) \qquad (8)$$

where $U_{art}(x)$ is the artificial potential field, $U_{att}(x)$ is the attractive potential field, and the $U_{rep}(x)$ is the repulsive potential field.

The potential forces are the gradients of the respective potential field,

$$F_{att} = -\nabla U_{att} \qquad (9)$$

$$F_{rep} = -\nabla U_{rep} \qquad (10)$$

When we consider multiple targets and obstacles (Fig 1), the final potential forces are the sum of attractive forces and repulsive forces. The attractive and repulsive potential forces are vectors, then the total force can be expressed as the sum of vectors.

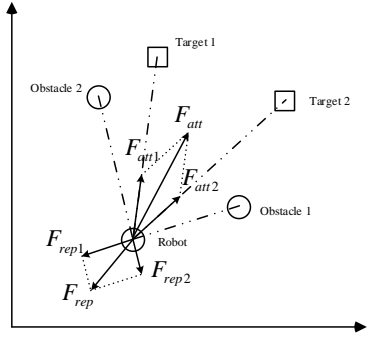

**Figure 1: Multiple targets and obstacles for artificial potential field.** The attractive force $\mathbf{F}_{att1}$ and $\mathbf{F}_{att2}$ produced by Target 1 and Target 2 are vectors. $\mathbf{F}_{att}$ is the vector sum of $\mathbf{F}_{att1}$ and $\mathbf{F}_{att2}$. Accordingly, $\mathbf{F}_{rep}$ is the vector sum of $\mathbf{F}_{rep1}$ and $\mathbf{F}_{rep2}$ produced by Obstacle 1 and Obstacle 2.

## 3. Combining Deep Reinforcement Learning and Safety Based Control

### 3.1. Methodology

In the field of cognitive science, there are two major learning paradigms, the empiricism and the speculation. Empiricism is a way of learning from historical experiences. Speculation is the way of logical thinking, which means taking measures by reasoning. The thinking process of humans contains both empiricism and speculation, which are interactive during the process.

The deep reinforcement learning method is just like learning from our past experiences. The safety based control, which contains artificial potential method and path tracking, is like the speculation and logical thinking. Using the deep reinforcement learning is efficient and can work well in a relative stable and familiar environment, but this method would be difficult to cover all scenarios. We combine the deep reinforcement learning and safety control to solve the problem.

### 3.2. Algorithm

We tackle the problem using perception sensor data, including vehicle speed, vehicle position on the road track and opponent vehicle distances (Loiacono et al., 2013). The input data can be divided into two parts. The features of opponent distances can be used for collision avoidance. Other parameters can be used for deep reinforcement learning and path tracking. Each of the three methods has its own acceleration and steering commands. We then balance the weight of these three action outputs,

$$\delta = \alpha \cdot \delta_l + \beta \cdot \delta_f + \gamma \cdot \delta_p \tag{11}$$

$$\tau = \alpha \cdot \tau_l + \beta \cdot \tau_f + \lambda \cdot \tau_p \tag{12}$$

$$\text{s.t.} \quad \alpha + \beta + \gamma = 1 \tag{13}$$

$\delta$, $\delta_l$, $\delta_f$ and $\delta_p$ respectively represents the final steering action, the learning policy steering action, the potential field steering action and the path tracking steering action. $\tau$, $\tau_l$, $\tau_f$ and $\tau_p$ respectively represents the final acceleration action, the learning policy acceleration action, the potential acceleration action and the path tracking acceleration action. $\alpha$, $\beta$ and $\gamma$ are respectively the weight parameters of the three methods.

### 3.2.1. Deep Reinforcement Learning

Firstly, we train the vehicle without opponents. The positive reward at each step is the velocity of the car projected along the track direction. We don't need to set negative rewards. The structure of DDPG is shown in Figure 2. The policy network is used to generate actions and the value function network is used to approximate the optimal Q-values. The input states are data from vehicle speed sensors, current engine speed, track sensors, wheel speed, track position and vehicle angle. After several hours training, we can use the policy to implement actions ($\delta_l$ and $\tau_l$) without opponents, which can also be applicable to other tracks because of the stable familiar environment.

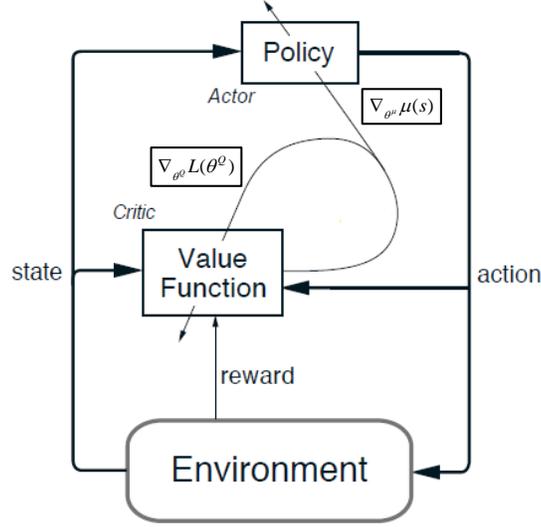

**Figure 2: Deep Deterministic Policy Gradient Policy Architecture.** The policy network is used to implement acceleration and steering demands. The input state parameters are partial sensor data. Then the action and state pairs are used in the critic network, which is the process of Q-learning. The actor-critic architecture can update the policy in the direction performance gradient $\nabla_\theta J(\pi_\theta)$.

By using the actor-critic structure, we can update critic by minimizing the loss,

$$L(\theta^Q) = E\left[(r(s_i,a_i) + \gamma Q(s_{i+1},\mu(s_{i+1});\theta^Q_{i-1}) - Q(s_i,a_i;\theta^Q))^2\right] \quad (14)$$

And we can update the actor policy by using the deterministic policy gradient,

$$\nabla_{\theta^\mu} J = E\left[\nabla_a Q^\mu(s,a) \cdot \nabla_{\theta^\mu}\mu(s)\right] \quad (15)$$

### 3.2.2. Artificial Potential Field Method

As for the artificial potential field, we only consider the repulsive potential field for collision avoidance. As is shown in the Figure 3, in the coordinate system of the ego vehicle, the $F_{rep}$ projected along the x-axis corresponds to the steering command and the force projected along the y-axis corresponds to the acceleration command. We assume the forces are continuous and only related to the distances,

$$F_{rep\_x} = -\sum_i \frac{1}{d_i^\eta} \cos \theta_i \tag{16}$$

$$F_{rep\_y} = -\sum_i \frac{1}{d_i^\eta} \sin \theta_i \tag{17}$$

where $\theta_i$ is the obstacle angle in the coordinate system of the ego vehicle, $d_i$ is the obstacle distance from the ego vehicle, and $\eta$ represents the power to be determined.

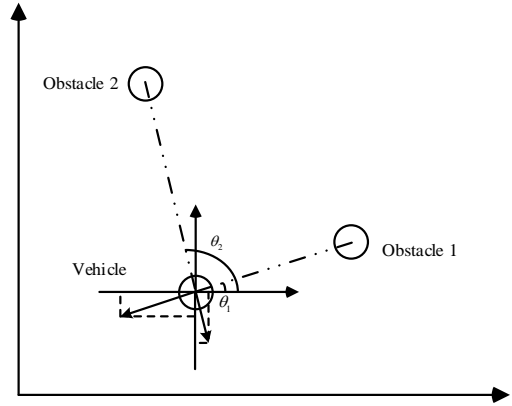

**Figure 3: Repulsive potential field forces in ego vehicle coordinate system.** The forces projected along the x-axis of ego vehicle correspond to the steering command. The forces projected along the y-axis correspond to the acceleration command.

Then the output actions are proportional to the potential field,

$$\delta_f = k_{fx} F_{rep\_x} \tag{18}$$

$$\tau_f = k_{fy} F_{rep\_y} \tag{19}$$

where $k_{fx}$ and $k_{fy}$ are respectively the proportional coefficients of $F_{rep\_x}$ and $F_{rep\_y}$.

**3.2.3. Path Tracking**

As for the path tracking function, we want the vehicle to drive along the central track of the road. The goal of path tracking is to minimize the angle between the car direction and the direction of track axis and shorten the distance between the vehicle centroid and the central road track (Kapania et al., 2015). The equation (20) represents the steering command with tracking error and heading error. We also tackle the acceleration command $\tau_p$ according to the steering command. The basic rule is to decrease the vehicle speed when the steering command is high enough.

$$\delta_p = \eta_1 \cdot \Delta \Psi + \eta_2 \cdot e \tag{20}$$

where $\Delta\Psi$ is the angle between the car direction and the direction of track axis, $e$ is the distance between the vehicle centroid and the central road track, and $\eta_1$ and $\eta_2$ are respectively their coefficients.

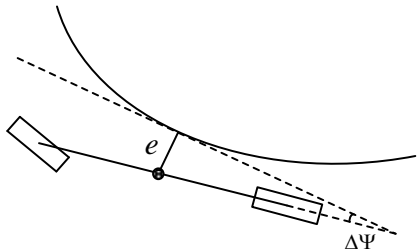

**Figure 4: Diagram showing tracking error $e$ and heading error $\Delta\Psi$.** The goal of path tracking is to minimize the tracking error and heading error to keep the vehicle drive along the track.

## 4. Experiments

### 4.1. Experiments Setup

We use TORCS platform (Loiacono et al., 2013) to implement our autonomous driving algorithm. First we train the policy network without opponents on GPU. The actor neural network consists of two hidden layers with 400 and 300 units respectively. The final output layer is a tanh layer to implement steering and acceleration commands. The policy network learning rate is $10^{-4}$. The critic neural network also consists of two hidden layers with 400 and 300 units with the learning rate $10^{-3}$. The discounted factor $\gamma$ is 0.99 and the training minibatch size is 64.

The input states for the actor-critic architecture are focus sensors, track sensors, vehicle speed, engine speed, wheel speed, track position and vehicle angle. The output actions are the steering commands and acceleration commands.

After the training, we combine the learning policy actions and safety based control actions. The parameters for the safety based control are shown in Table 1. The input states for the repulsive potential field are the 36 opponent distances. The path tracking method uses the angle error and track position to calculate the actions.

**Table 1:** Parameters for the artificial potential field and path tracking

| Symbol | Value | Symbol | Value |
|---|---|---|---|
| $k_{fx}$ | 20 | $\eta_2$ | 2 |
| $k_{fy}$ | 10 | $\alpha$ | 0.4 |
| $\eta$ | 1.5 | $\beta$ | 0.3 |
| $\eta_1$ | 3.18 | $\gamma$ | 0.3 |

### 4.2. Results

We first train the driving policy using DDPG algorithm without opponents on GPU. The average Q-value of the actor-critic structure has increased gradually (Fig 5). During the training process, we divide the reward by 150 to limit the one-step reward to $[0, 2]$. After approximate 13 hours training, the average Q-value reaches approximate 110. We then use the policy for autonomous driving. The vehicle could perform well using the trained policy network.

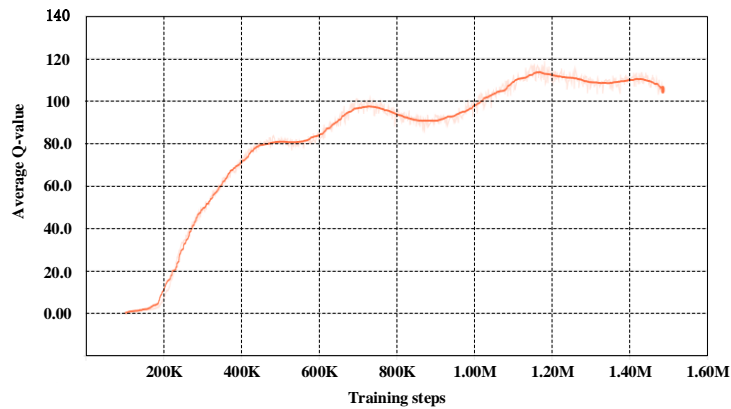

**Figure 5: Average Q-value during the training process.** After 13 hours, the average value reaches 110, we then used the policy network for autonomous driving without vehicles around.

We then combine the DDPG policy network and safety based control. The ratio coefficients for DDPG, Path Tracing and Artificial Potential Field (APF) method are 0.4, 0.3 and 0.3. Figure 6(a)~6(d) show 4 typical scenarios using the combined algorithm and Figure 6(d)~6(e) show the respective steering commands and acceleration commands by DDPG, Path Tracking and APF.

Figure 6(a) shows the vehicle runs along the curve. The DDPG algorithm outputs the major steering command and the APF commands for steering and acceleration are 0 because no vehicle around is detected. Figure 6(b) and Figure 6(c) show that there is one opponent around and APF commands output corresponding actions. The opponent distance in Figure 6(c) is shorter than Figure 6(b), so the APF commands play the major parts in Figure 6(c). Figure 6(d) shows the scenario in which the vehicle runs along the curve with two vehicles around. The ego vehicle is far from the track so the Path Tracking steering command is higher than two other methods.

**5. Conclusion**

In this paper, we combine the deep reinforcement learning and safety based control, including artificial potential field and path tracking for autonomous driving. We first use the DDPG algorithm to get the driving policy using partial state inputs and then combine the policy network and safety based control to avoid collision and drive along the track. Experiments show that the three algorithms coordinate well in the TORCS environment.

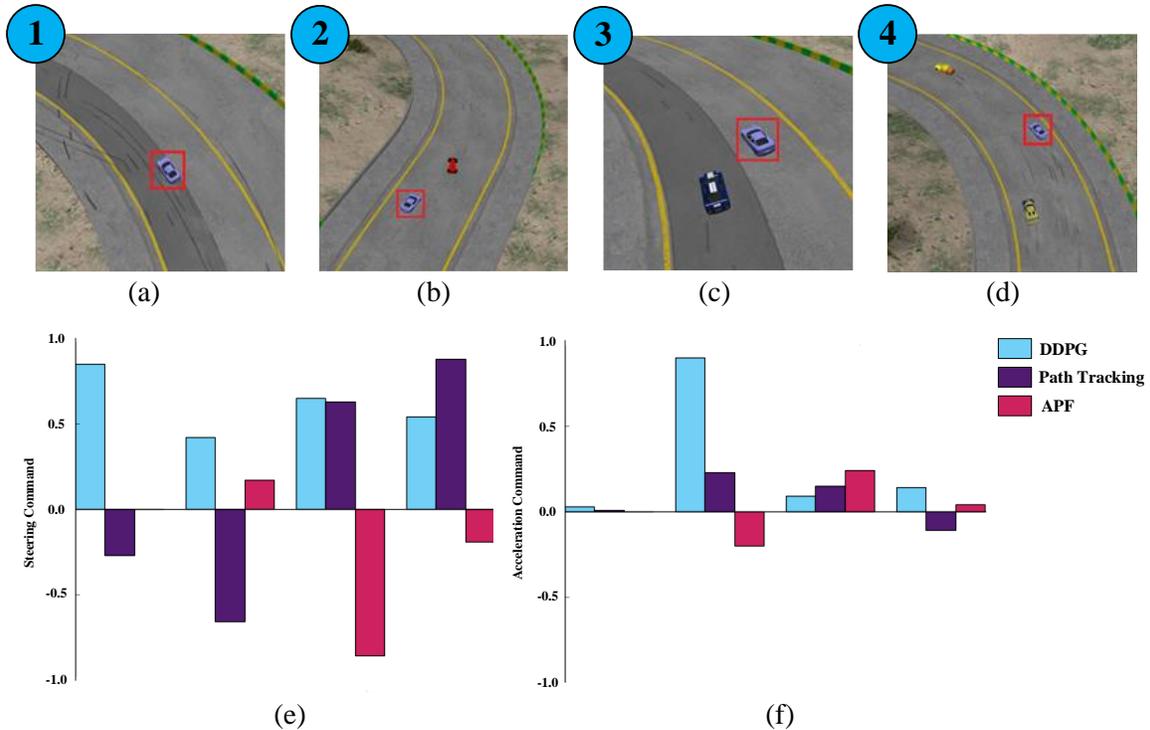

**Figure 6: Typical driving scenarios and corresponding commands.** The positive steering command represents turning left and vice cersa. The negative acceleration command represents braking control. The blue race car with red box is the ego vehicle. (a), Driving along the curve with no opponent around. (b), One opponent vehicle in front. (c), One opponent in bottom left. (d), Two opponents around while driving along the curve. (e), The steering commands in the four typical scenarios using DDPG, Path Tracking and APF. (f), The accelearation commands in the four typical scenarios using DDPG, Path Taracking and APF.